\documentclass{article}
\usepackage{ijcai26}

\usepackage{times}
\usepackage{soul}
\usepackage{url}
\usepackage[hidelinks]{hyperref}
\usepackage[utf8]{inputenc}
\usepackage[small]{caption}
\usepackage{graphicx}
\usepackage{amsmath}
\usepackage{amsthm}
\usepackage{adjustbox}
\usepackage{booktabs}
\usepackage{algorithm}
\usepackage{algorithmic}
\usepackage[switch]{lineno}

\title{Derain-Agent: A Plug-and-Play Agent Framework for Rainy Image Restoration}

\author{
    Zhaocheng Yu,
    Xiang Chen,
    Runzhe Li,
    Zihan Geng,
    Guanglu Sun,
    Haipeng Li,
    Kui Jiang\thanks{Corresponding Author} \\
    \emails
    \{yuzhaocheng, lirunzhe\}@stu.hit.edu.cn,
chenxiang@njust.edu.cn,
geng.zihan@sz.tsinghua.edu.cn,
sunguanglu@hrbust.edu.cn,
lihaipeng@gml.ac.cn,
jiangkui@hit.edu.cn
}

\usepackage{amssymb}
\begin{document}

\maketitle

\begin{abstract}

While deep learning has advanced single-image deraining, existing models suffer from a fundamental limitation: they employ a static inference paradigm that fails to adapt to the complex, coupled degradations (\emph{e.g.}, noise artifacts, blur, and color deviation) of real-world rain. Consequently, restored images often exhibit residual artifacts and inconsistent perceptual quality. In this work, we present \textbf{Derain-Agent}, a plug-and-play refinement framework that transitions deraining from static processing to dynamic, agent-based restoration. Derain-Agent equips a base deraining model with two core capabilities: 1) a \textbf{Planning Network} that intelligently schedules an optimal sequence of restoration tools for each instance, and 2) a \textbf{Strength Modulation} mechanism that applies these tools with spatially adaptive intensity. This design enables precise, region-specific correction of residual errors without the prohibitive cost of iterative search. Our method demonstrates strong generalization, consistently boosting the performance of state-of-the-art deraining models on both synthetic and real-world benchmarks.
\end{abstract}

\section{Introduction}

Rain severely degrades image quality, 
compromising the reliability of vision systems in safety‑critical domains such as autonomous driving and video surveillance. 
Consequently, single‑image deraining, 
the task of reconstructing clear images from rain-degraded inputs, is essential for 
enabling robust outdoor perception systems~\cite{dang2024adaptive,dang2023efficient}.
\begin{figure}[h]
\centering
\includegraphics[width=\linewidth]{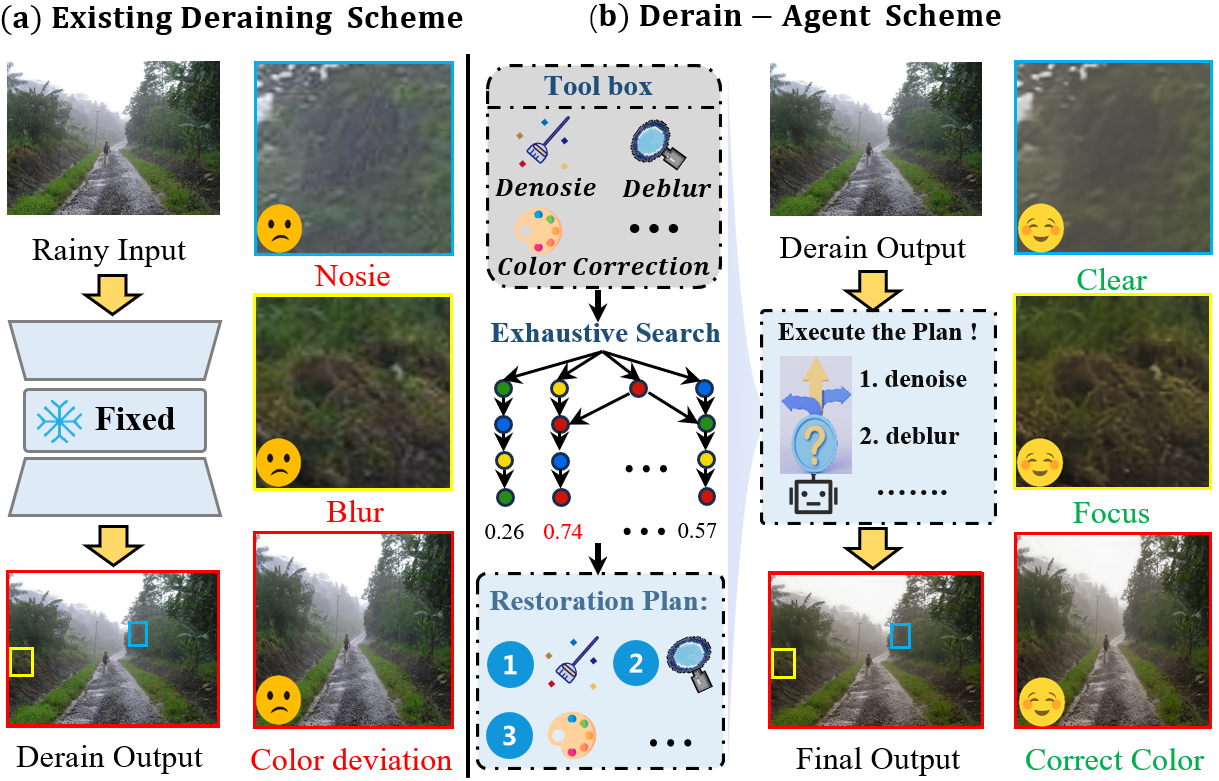}
\vspace{-4mm}
\caption{Preliminary findings: a) static deraining leaves coupled residual degradations; b) agent-based adaptive refinement markedly improves visual quality.}
\vspace{-3mm}
\label{moti}
\end{figure}
While deep learning~\cite{lecun2015deep} has driven significant progress in this area, with modern approaches achieving impressive performance on standard benchmarks~\cite{wang2020dcsfn,li2024fouriermamba,xiao2022image}, 
a notable gap remains between benchmark results and real-world applicability. 
Under real-world conditions, these models frequently produce results of insufficient perceptual quality. 
A primary reason for this shortfall is that 
prevailing methods ~\cite{cui2023selective,chen2021pre} are predominantly optimized for synthetic benchmarks, 
whereas real-world rain involves coupled degradations (\emph{noise, motion, and splatter}) and complex global photometric disturbances. 
This often leads to residual artifacts, blur, and color deviation in preliminary deraining output.
Recurrent and cascaded architectures~\cite{ren2019progressive,jiang2022danet,jiang2022magic} represent a progressive effort to improve both \emph{fidelity} and \emph{textural plausibility} by jointly addressing rain-streak removal and detail recovery. 
Yet they still operate with static inference: the trained architecture executes a fixed processing routine for all inputs, lacking dynamic diagnosis and sample-adaptive adjustment to varying degradation conditions.
Thus, the fundamental bottleneck lies not in network capacity, but in the static inference paradigm—the inability to dynamically diagnose and adaptively treat the diverse, residual degradations present in each case.


The emerging paradigm of agent-based image restoration~\cite{restoreagent2024,zhou2025q,agenticir2025} offers a promising direction to address this limitation. We posit that an agent-based paradigm is inherently suited for this challenge, enabling instance-level, adaptive, and dynamic refinement of deraining outputs. 
Preliminary analysis supports this: as shown in Fig.~\ref{moti}, applying an optimal sequence of enhancement tools (found via brute-force search) to a preliminary deraining result yields marked perceptual improvement. 
However, 
this brute-force approach has two critical drawbacks: 1) the prohibitive computational cost of exhaustive search renders it impractical, 
and 2) 
the use of fixed, preset tool strengths lacks the flexibility needed to handle complex, spatially varying, and coupled degradations. 

To address these issues, we propose Derain-Agent, a plug-and-play agent refinement framework for the adaptive refinement of single-image deraining outputs. 
Given the output of any base deraining model, Derain-Agent operates in two core stages. First, to eliminate the need for exhaustive search, we introduce a lightweight Planning Network that learns to predict an effective, instance-specific sequence of tools directly from the input. Second, to manage coupled degradations, we design a Strength Modulation module that generates a spatially adaptive strength map for each tool, allowing for fine-grained, region-specific enhancement. 
This fine-grained control enables the simultaneous suppression of  
residual artifacts 
and preservation of structural details and color fidelity. 
In general, the contributions of this work can be summarized as follows.
\begin{itemize}
    \item We introduce Derain-Agent, 
    a novel plug-and-play agent framework that dynamically refines deraining outputs. 
    It uniquely integrates \emph{adaptive tool scheduling} with \emph{spatially-varying strength modulation} to perform progressive, instance-specific restoration. 
    \item 
    We design two core learnable components: a lightweight \emph{Planning Network} that efficiently predicts optimal tool sequences, and a \emph{Strength Modulation Module} that generates spatially adaptive maps for fine-grained control, together overcoming the limitations of brute-force search and fixed parameters.
    \item Through extensive experiments, we demonstrate that Derain-Agent serves as a universal enhancer, boosting the performance of diverse deraining backbones on both synthetic and real-world datasets, achieving state-of-the-art \emph{cross-domain generalization}. 
    
\end{itemize}

\section{Related Work}
\subsection{Single Image Deraining}
Early single-image deraining methods relied predominantly on hand-crafted priors and model-based optimization.
While effective in constrained or synthetic settings, 
these approaches often struggle with the diverse and complex rain patterns observed in real-world imagery.
The advent of deep learning has substantially improved deraining performance, 
leading to a wide spectrum of architectures, including feed-forward CNN and Transformer models, recurrent or progressive refinement designs, and more recent generative formulations.
Recent progress has been driven by 
advances in stronger representations and architectural designs, such as frequency-aware modeling, implicit restoration formulations \cite{gao2024efficient,chen2024bidirectional}, and efficient state-space models for long-range dependency modeling in restoration tasks \cite{liu2024vmamba,zhen2024freqmamba}.
Nevertheless, bridging the synthetic-to-real gap remains a key challenge, as real rainfall often involves coupled degradations and imaging effects beyond simple streak patterns.
Despite these advances, a persistent limitation remains: most derainers 
exhibit a fixed restoration behavior during inference 
and lack the adaptability to address 
diverse, coupled, and instance-specific degradations in real rainy scenes. This ultimately restricts robustness and real-world generalization.
Our work addresses this limitation 
by introducing an agent-based, plug-and-play refinement stage that \emph{adaptively refines} deraining outputs. 
\subsection{Agent-based Vision Systems}
Recent studies have increasingly formulated low-level image restoration as a sequential decision-making problem, where an agent diagnoses degradations and composes a sequence of operators to progressively refine perceptual quality. For instance, AgenticIR demonstrates a perception–scheduling–execution–reflection loop for 
complex compound degradations (\emph{e.g.}, rain with haze or noise) \cite{agenticir2025}. RestoreAgent leverages MLLMs to analyze degradations and dynamically select restoration models 
according to user-defined objectives \cite{restoreagent2024}. To reduce planning cost, MAIR introduces structured real-world degradation priors and a hierarchical multi-agent architecture \cite{mair2025}. Quality-aware control has been further explored by incorporating no-reference IQA feedback into sequential decision-making, as in 4KAgent \cite{4kagent2025} and Q-Agent \cite{zhou2025q}. Meanwhile, HybridAgent highlights the issue of error accumulation in stepwise pipelines and mitigates it 
through collaborative agent designs \cite{hybridagent2024}.
Inspired by this agentic restoration perspective, we 
frame rainy-scene enhancement as an instance-adaptive refinement problem built upon an initial deraining output.
In real images, 
the removal of rain streaks alone is often insufficient, as rainfall 
frequently entails coupled degradations whose presence and severity vary per scene.
An agentic formulation naturally enables the system to assess residual degradations 
post-deraining and conditionally schedule subsequent enhancement actions.
This insight motivates our plug-and-play refinement framework, which applies lightweight test-time diagnosis and adaptive tool scheduling to improve perceptual quality without altering the 
base deraining model.

\section{Method}
This section details the proposed Derain-Agent, including its problem formulation, overall pipeline, and the design of its perception and execution components. 

\begin{figure*}[!htpb]
  \centering
  \includegraphics[width=\textwidth]{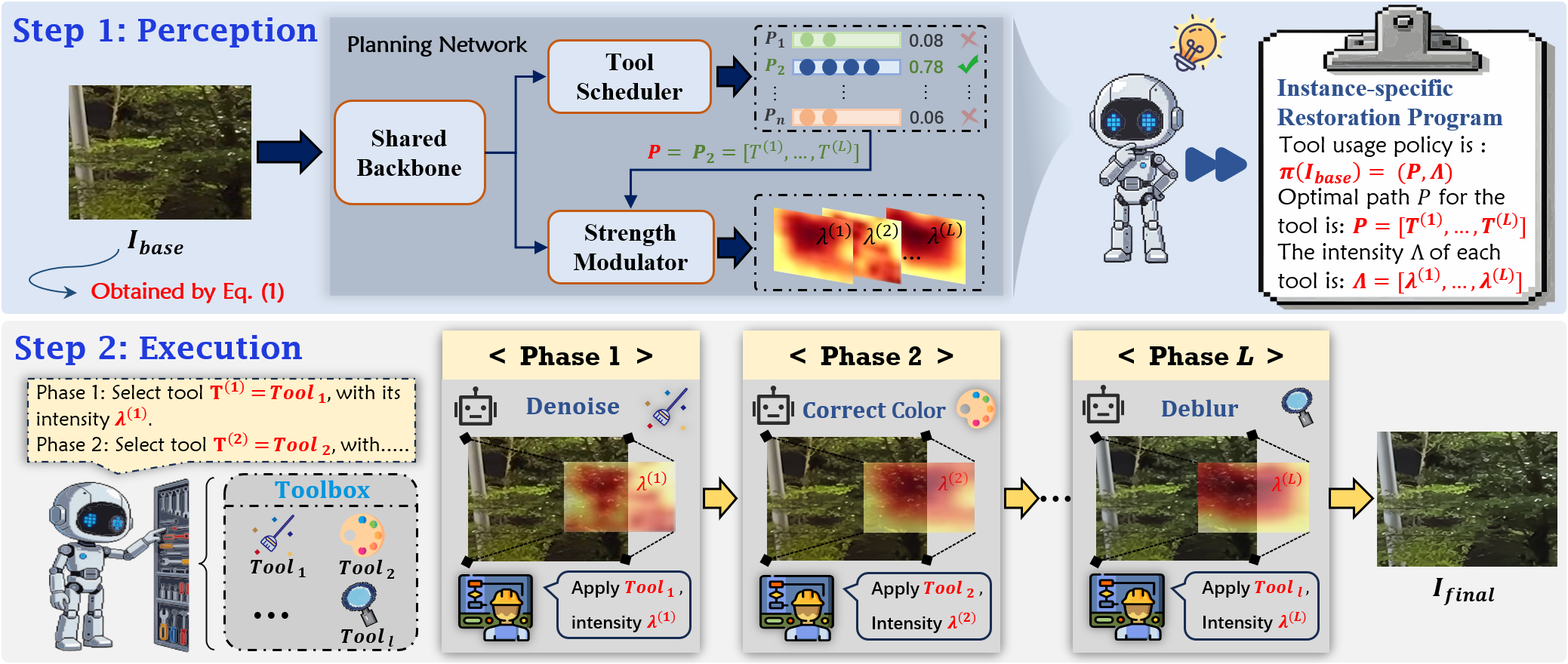}
  \vspace{-2mm}

  \caption{Overview of the Derain-Agent framework. Given a preliminary deraining output $I_{\text{base}}$, the perception stage analyzes residual degradations to produce an executable restoration program $\Pi=(P, \Lambda)$. The execution stage then applies the scheduled tools $T^{(\ell)}$ sequentially, modulated by pixel-wise strength maps $\lambda^{(\ell)}$, to yield the final enhanced image $I_{\text{final}}$.}
  \label{pipeline}
\end{figure*}

\subsection{Problem Formulation}
Conventional single-image deraining methods learn 
a fixed mapping 
$G$ from a rainy input image $I_{rain}$ to a restored output:
\begin{equation}
I_{\text{base}} = G(I_{rain}).
\end{equation}
%
%

However, in real-world scenes, rainfall is rarely an isolated degradation. 
Beyond visible rain streaks, 
it often introduces coupled perturbations such as noise artifacts, background blur, and global color distortion. 
As the presence and severity of these residual degradations vary 
per instance, the preliminary deraining result $I_{\text{base}}$ frequently 
exhibits inconsistent perceptual quality across diverse scenes.

To address this limitation, we 
formulate post-deraining enhancement as an  instance-adaptive, tool-based refinement process. 
Given a preliminary result $I_{\text{base}}$, the Derain-Agent performs 
a perceptual analysis of residual degradations and outputs  a restoration program $\Pi(I_{\text{base}})$. This program is formalized as a tuple: 
\begin{equation}
\Pi(I_{\text{base}}) = (P, \Lambda),
\label{Π}
\end{equation}
where
\begin{equation}
P = [T^{(1)}, \dots, T^{(L)}], \quad T^{(\ell)} \in \mathcal{T},
\label{P}
\end{equation}
denotes a tool path—an ordered sequence of operators from a predefined toolbox $\mathcal{T}$—determined by a scheduling policy. The term 
\begin{equation}
\Lambda = [\lambda^{(1)}, \dots, \lambda^{(L)}],\quad \lambda^{(\ell)} \in [0,1]
\label{λ}
\end{equation}
represents spatially varying strength maps that modulate the execution intensity of each scheduled tool.

Starting from $I^{(0)} = I_{\text{base}}$, the execution process is carried out sequentially as
\begin{equation}
I^{(\ell)} = T^{(\ell)}\big(I^{(\ell-1)}, \lambda^{(\ell)}\big), \quad \ell = 1, \dots, L,
\label{step}
\end{equation}
yielding the final enhanced output $I^{(L)}$.

\subsection{Pipeline Overview}
\textbf{Overall Architecture.} 
As shown in Figure~\ref{pipeline}, Derain-Agent 
employs a perception--execution architecture to enhance any preliminary deraining output $I_{\text{base}}$.

\textbf{(1) Perception Stage.} The input $I_{\text{base}}$ is first 
processed by a shared feature backbone to obtain a representation indicative of residual degradations. 
This representation is fed into two lightweight planners: a \textit{Tool Scheduler} that predicts the optimal ordered path $P$, and a \textit{Strength Modulator} that predicts the corresponding pixel-wise strength maps $\Lambda$. Together, they constitute the executable restoration program $\Pi(I_{\text{base}})$. 

\textbf{(2) Execution Stage.} The agent executes the restoration program $\Pi$ sequentially. At step $\ell$, it invokes tool $T^{(\ell)}$ and applies it to the intermediate image $I^{(\ell-1)}$, using the strength map $\lambda^{(\ell)}$ for spatially adaptive modulation to produce $I^{(\ell)}$. After $L$ steps, the pipeline outputs the final enhanced image $I_{\text{final}}$.
\textbf{Model Optimization.}
The model is optimized in two distinct stages to first learn the tool-scheduling policy and then the strength modulation parameters.

\textbf{Stage 1: Path Scheduling.} We jointly train the shared backbone and the tool-path scheduling head using a cross-entropy loss:
\begin{equation}
\mathcal{L}_{\text{path}}=-\sum_{c=1}^{C} y_c \log(p_c),
\end{equation}
where $C$ is the total number of 
predefined path categories, $y$ is the one-hot vector of the ground-truth path, and $p_c$ denotes the predicted probability for category $c$.
The ground-truth path for each training sample is generated offline by 
first generating the preliminary result $I_{\text{base}}$ with the base derainer and then performing exhaustive search over the toolbox to select the 
tool sequence that yields the best performance under paired supervision. 

\textbf{Stage 2: Strength Modulation.} With the backbone and path scheduler frozen, we optimize only the strength modulation head. The final output $I_{\text{final}}$ is supervised by the paired clean image $I_{\text{gt}}$ using a composite reconstruction loss:
\begin{equation}
\mathcal{L}_{\text{recon}}=\|I_{\text{final}}-I_{\text{gt}}\|_1+\mu\cdot (1-\text{SSIM}(I_{\text{final}},I_{\text{gt}})),
\end{equation}
where $\mu=0.1$ balances the loss terms.

\subsection{Perception}
The perception stage 
is implemented by a Planning Network, which 
maps the preliminary derained result $I_{\text{base}}$ 
to an executable restoration program $\Pi(I_{\text{base}})=(P,\Lambda)$.
It comprises a shared backbone for feature extraction and two lightweight prediction heads: a tool scheduler that 
outputs the discrete execution order $P$, and a strength modulator that predicts the pixel-wise 
modulation maps $\Lambda$ for 
spatially adaptive control.

\textbf{Shared Backbone.}
We 
employ a ResNet34~\cite{he2016deep} as the shared backbone to extract a compact feature representation $F \in \mathbb{R}^{(c,h,w)}$ from $I_{\text{base}}$.
$F$ serves as the common input for both downstream planning heads. 

\textbf{Tool Scheduler.}
This head is formulated as a lightweight classifier. 
The feature map $F$ 
is first globally average-pooled and flattened, 
and then passed through two fully connected layers to produce a logit vector over $C$ candidate path categories.
The 
scheduled path $P$ is obtained by selecting the 
category with the highest probability.
Each category corresponds to a predefined, 
ordered tool execution sequence (\emph{e.g.}, denoising followed by deblurring), which 
is executed in the subsequent execution stage.


\textbf{Strength Modulator.}
Conditioned on the scheduled path $P$ and the shared feature map $F$, 
this head predicts a set of pixel-wise strength maps $\Lambda=[\lambda^{(1)},\ldots,\lambda^{(L)}]$, one for each tool in the path. 
Specifically, 
the discrete path $P$ 
is encoded as a one-hot vector and projected into a compact conditional embedding $E_P$.
This embedding is spatially tiled and concatenated with $F$ along the channel dimension to form a path-conditioned feature.
This fused feature is then processed by three consecutive convolutional layers to produce final strength maps $\Lambda$, where $\lambda^{(\ell)}$ corresponds to the tool $T^{(\ell)}$ at step $\ell$ along the planned path.
During execution, each map $\lambda^{(\ell)}$ acts as a pixel-wise scaling factor  to modulate the intensity 
of the corresponding tool $T^{(\ell)}$. 

\subsection{Execution}
Given the executable program $\Pi(I_{\text{base}})=(P,\Lambda)$ generated by the Planning Network, 
Derain-Agent performs step-wise refinement by sequentially executing the scheduled tools, as outlined in Algorithm~\ref{alg:execution}.

Starting from the initial state $I^{(0)} = I_{\text{base}}$, at each step $\ell$ the agent selects the tool $T^{(\ell)}$ from the planned path $P = [T^{(1)}, \ldots, T^{(L)}]$ and applies it to the current intermediate image $I^{(\ell-1)}$. The effect of the tool is then modulated in a spatially adaptive manner by the corresponding pixel-wise strength map $\lambda^{(\ell)}$ (see Figure~\ref{pipeline}), producing the updated result $I^{(\ell)}$. After $L$ steps, the pipeline outputs the final enhanced image $I_{\text{final}} = I^{(L)}$.
The strength-controlled execution is implemented via a residual scaling mechanism: 
\begin{equation}
I^{(\ell)} = I^{(\ell-1)} + \lambda^{(\ell)} \odot \big(f_{T^{(\ell)}}(I^{(\ell-1)}) - I^{(\ell-1)}\big),
\label{eq:exec}
\end{equation}
where $f_{T^{(\ell)}}(\cdot)$ denotes the standard operator 
corresponding to tool $T^{(\ell)}$, and $\odot$ represents element-wise multiplication.

\begin{algorithm}[t]
\caption{Execute the Restoration Program}
\label{alg:execution}
\begin{algorithmic}[1]
\STATE \textbf{Input:} preliminary derain output $I_{\text{base}}$; restoration program $\Pi(I_{\text{base}})=(P,\Lambda)$ with
$P=[T^{(1)},\ldots,T^{(L)}]$ and $\Lambda=[\lambda^{(1)},\ldots,\lambda^{(L)}]$; frozen tools $\{f_T(\cdot)\}_{T\in\mathcal{T}}$
\STATE \textbf{Output:} enhanced image $I_{\text{final}}$
\STATE $I^{(0)} \leftarrow I_{\text{base}}$
\FOR{$\ell = 1$ \textbf{to} $L$}
    \STATE $I_{\text{tool}} \leftarrow f_{T^{(\ell)}}(I^{(\ell-1)})$ \hfill \COMMENT{apply the scheduled tool}
    \STATE $I^{(\ell)} \leftarrow I^{(\ell-1)} + \lambda^{(\ell)} \odot \big(I_{\text{tool}} - I^{(\ell-1)}\big)$
    \hfill \COMMENT{strength modulation}
\ENDFOR
\STATE $I_{\text{final}} \leftarrow I^{(L)}$
\end{algorithmic}
\end{algorithm}
\vspace{-2mm}

\begin{table*}[t]
\centering
\caption{Performance (PSNR / SSIM) comparison on the LHP-Rain dataset.
Baseline derainers are trained on different synthetic datasets and enhanced using their corresponding Derain-Agents.}\vspace{-2mm}
\label{table1}
\begin{adjustbox}{width=\linewidth}
\setlength{\tabcolsep}{7pt}
\renewcommand{\arraystretch}{1.15}
\begin{tabular}{l c c c c c c c}
\toprule
\textbf{Rain13K} → \textbf{LHP-Rain}&
\textbf{Sfnet} &
\textbf{Restormer} &
\textbf{MFDNet} &
\textbf{NeRD-Rain} &
\textbf{FADformer} &
\textbf{FreqMamba}&
\textbf{Average}\\
\midrule


Baseline derainers
& 31.30 / 0.8906 & 31.59 / 0.8835 & 31.66/ 0.8855 & 30.83 / 0.8854 & 31.27 / 0.8969&30.12 / 0.8937 &31.12 / 0.8892\\
+ Derain-Agent (ours)
& 32.03 / 0.9080 & 32.34 / 0.9074 & 32.46 / 0.9079 & 31.62 / 0.9083 &31.92 / 0.9074  &31.22 / 0.9034&31.93 / 0.9070 \\

\midrule
\textbf{Rain200H}→ \textbf{LHP-Rain} &
\textbf{IDT} &
\textbf{DRSformer} &
\textbf{MSDT} &
\textbf{NeRD-Rain} &
\textbf{FADformer} &
\textbf{TransMamba}&
\textbf{Average}
 \\
\midrule

Baseline derainers
& 30.34 / 0.8741 &  30.10 / 0.8640 &26.72 / 0.8050&27.69 / 0.8517& 22.57 / 0.8198 & 29.82 / 0.8668 &27.87 / 0.8469 \\
+ Derain-Agent (ours)
& 31.14 / 0.9013 & 31.03 / 0.8931 &27.63 / 0.8294 &28.61 / 0.8684 & 25.19 / 0.8275 & 30.78 / 0.8917 &29.06 / 0.8685\\

\bottomrule
\end{tabular}
\end{adjustbox}
\vspace{-2mm}
\end{table*}

\begin{table*}[t]
\centering
\caption{ Performance (PSNR / SSIM) comparison on the LHP-Rain dataset using a unified Derain-Agent.
All baseline derainers are trained on Rain200H.}\vspace{-2mm}
\label{table2}
\begin{adjustbox}{width=\linewidth}
\setlength{\tabcolsep}{6pt}
\begin{tabular}{lccccccc}
\toprule
\textbf{Method}  & \textbf{IDT} & \textbf{DRSformer} & \textbf{MSDT}&  \textbf{NeRD-Rain} & \textbf{FADformer} & \textbf{TransMamba} & \textbf{Average}\\
\midrule
Baseline derainers
& 30.34 / 0.8741
& 30.10 / 0.8640
& 26.72 / 0.8050
& 27.69 / 0.8517
& 22.57 / 0.8198
& 29.82 / 0.8668
& 27.87 / 0.8467 \\

+ Derain-Agent (Ours)
& 30.95 / 0.9017
& 30.86 / 0.8933
& 27.33 / 0.8349
& 28.25 / 0.8796
& 24.01 / 0.8286
& 30.62 / 0.8944
& 28.67 / 0.8720\\
\bottomrule
\end{tabular}
\end{adjustbox}
\vspace{-2mm}
\end{table*}
\section{Experiment}

This section describes the implementation, training protocol, and evaluation of Derain‑Agent. Derain-Agent is interfaced with multiple deraining backbones pretrained on synthetic data, and is evaluated on real-world benchmarks under both paired and unpaired settings using full-reference and no-reference image quality metrics.

\subsection{Implementation Details}
\textbf{Datasets.}
We evaluate Derain-Agent 
on a combination of synthetic and real-world rain benchmarks, including LHP-Rain~\cite{guo2023sky}, RE-Rain~\cite{chen2023towards}, Rain13K~\cite{zamir2021multi}, and Rain200H~\cite{yang2017deep}, which cover diverse rain patterns, scenes, and distributions.
The base derainers are trained on the synthetic datasets Rain13K and Rain200H, which represent diverse synthetic rain appearances and heavy-rain degradation, respectively.

For paired real-scene evaluation, we use LHP-Rain~\cite{guo2023sky}, a high-resolution benchmark that provides aligned rainy/clean image pairs captured under  real rain conditions with coupled imaging effects.
Following the standard protocol, we train Derain-Agent using approximately 10{,}000 rainy/clean pairs randomly sampled from the LHP-Rain training split, and we report quantitative results on the official test set (1{,}000 images from 300 sequences).
For unpaired real-world evaluation where ground truth is unavailable, we use RE-Rain~\cite{chen2023towards}, which contains 300 real rainy images, and report no-reference results together with qualitative comparisons in unconstrained real scenes.


\begin{figure*}[!htpb]
  \centering
  \includegraphics[width=\textwidth]{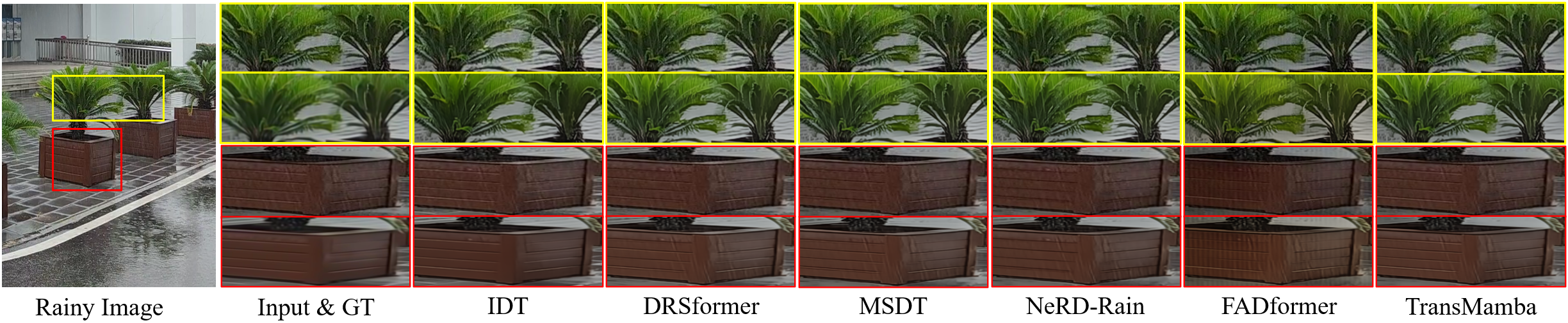}\vspace{-2mm}

  \caption{Visual results on LHP-Rain. Top: Baseline derainers; Bottom: Derain-Agent enhanced. Please zoom in for a better view.}
  \label{One_to_one}\vspace{-2mm}
\end{figure*}

\begin{figure*}[!htpb]
  \centering
  \includegraphics[width=\textwidth]{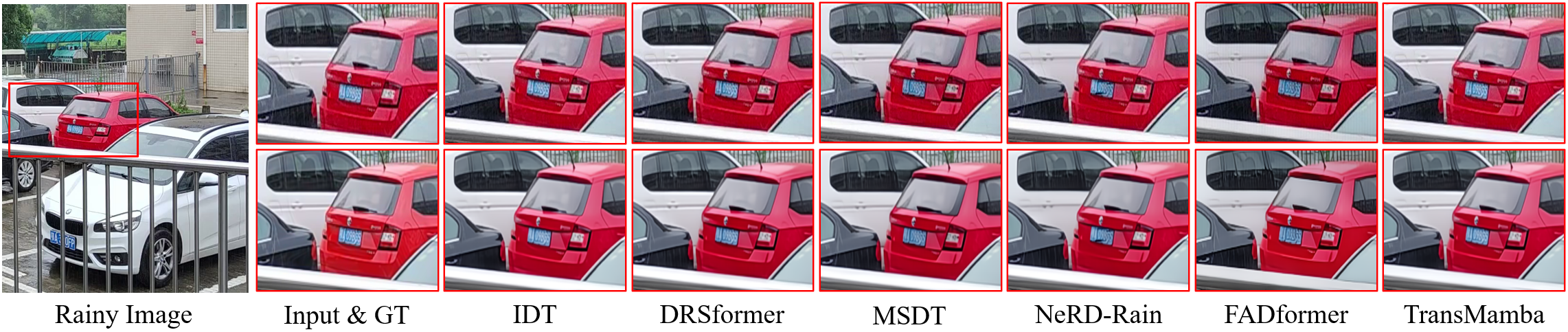}\vspace{-2mm}

  \caption{Visual results on LHP-Rain. Top: Baseline derainers; Bottom: Derain-Agent enhanced. Please zoom in for a better view.}
  \label{all_to_one}\vspace{-4mm}
\end{figure*}

\begin{table*}[t]
\centering
\caption{No-reference evaluation (BRISQUE / NIQE) on the unpaired RE-Rain dataset.
All baseline derainers are trained on Rain200H, and results are reported with and without the unified Derain-Agent. Lower is better.}\vspace{-2mm}
\label{tab:rerain}
\begin{adjustbox}{width=\linewidth}
\setlength{\tabcolsep}{6pt}
\begin{tabular}{lccccccc}
\toprule
\textbf{Method}  & \textbf{IDT} & \textbf{DRSformer} & \textbf{MSDT}&  \textbf{NeRD-Rain} & \textbf{FADformer} & \textbf{TransMamba} & \textbf{Average} \\
\midrule
Baseline derainers 
&23.704 / 3.918&  22.857 / 3.875 &20.128 / 3.793 & 21.693 / 3.862 & 21.670 / 3.980 & 23.420 / 3.844& 22.245 / 3.878 \\
+ Derain-Agent (ours) 
&22.148 / 3.806 & 20.059 / 3.780 & 19.693 / 3.609  &  19.866 / 3.793 &  19.725 / 3.917 &  21.070 / 3.761 & 20.426 / 3.777 \\
\bottomrule
\end{tabular}\vspace{-4mm}
\end{adjustbox}
\end{table*}
\textbf{Metrics.}
For datasets with ground truth, we report PSNR~\cite{huynh2008scope} and SSIM~\cite{wang2004image}.
For unpaired real-world images without ground truth, we report no-reference image quality assessment metrics, including BRISQUE~\cite{mittal2012no} and NIQE~\cite{mittal2012making}, to assess perceptual quality.

\textbf{Experimental Settings.}
The toolbox comprises three frozen tools for distinct degradations: SCUNet~\cite{zhang2023practical} for denoising, WB LUTs~\cite{manne2024wb} for color correction, and Restormer~\cite{zamir2022restormer} for deblurring.
Therefore, the number of exhaustive paths is C = 16 (including the case of using no tools).
We adopt a two-stage training scheme to decouple tool-path scheduling from execution strength learning.
In Stage~1, 
we train the shared backbone and the tool-path scheduling head with an initial learning rate of $2 \times 10^{-4}$.
In Stage~2, 
we freeze the backbone and the scheduling head, and optimize only the intensity modulation head with an initial learning rate of $1 \times 10^{-4}$, which stabilizes the learning of execution strength under fixed tool paths. 
Both stages use the Adam optimizer with a cosine-annealing schedule that decays the learning rate to $1 \times 10^{-6}$.
Input images are center-cropped to $256 \times 256$ with a batch size of 32.
All experiments are implemented in PyTorch and conducted on a single NVIDIA 3090 GPU.

\subsection{Results and Analysis}
\textbf{Specific Agents for Cross-Dataset Generalization.}
Table~\ref{table1} presents the cross-dataset evaluation on the LHP-Rain benchmark.
Derain-Agent consistently enhances performance across all backbones, yielding an average PSNR gain of \textbf{0.81 dB} when trained on Rain13K.
Notably, under the more challenging Rain200H setting—characterized by a larger domain gap and weaker baseline performance for models like MSDT and FADformer—Derain-Agent achieves an even higher average gain of \textbf{1.19 dB}. This validates the robustness of our instance-adaptive refinement strategy.
As illustrated in Figure~\ref{One_to_one}, Derain-Agent effectively suppresses residual noise and blur persisting in the derained outputs.
This advantage stems from its ability to explicitly diagnose and correct coupled residual degradations via dynamic tool scheduling, rather than relying on a fixed, one-size-fits-all restoration mapping.

\begin{figure*}[!htpb]
  \centering
  \includegraphics[width=\textwidth]{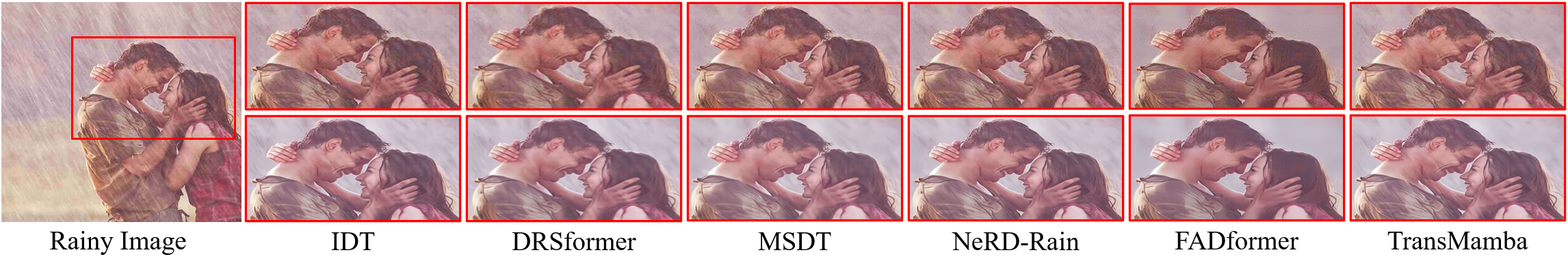}\vspace{-2mm}

  \caption{Visual results on RE-Rain. Top: Baseline derainers; Bottom: Derain-Agent enhanced. Please zoom in for a better view.}
  \label{RE-Rain}\vspace{-4mm}
\end{figure*}

\textbf{Unified Agent for Cross-Algorithm Generalization.} 
To assess the universality of our approach, we train a single unified Derain-Agent and apply it to multiple backbones in a plug-and-play manner.
As shown in Table~\ref{table2}, this unified agent consistently improves all evaluated derainers on LHP-Rain, indicating that the learned refinement policy transfers effectively across diverse architectures without requiring model-specific tuning. 
Qualitatively, Figure~\ref{all_to_one} demonstrates that Derain-Agent recovers clearer structures and reduces residual artifacts.
Furthermore, on the unpaired real-world RE-Rain dataset (Table~\ref{tab:rerain}), Derain-Agent achieves consistently lower NIQE and BRISQUE scores.
Visual comparisons in Figure~\ref{RE-Rain} confirm that our method produces cleaner results with more natural color fidelity.
Collectively, these results demonstrate that Derain-Agent provides robust, algorithm-agnostic refinement suitable for complex real-world conditions.

\textbf{Model Complexity.}
Table~\ref{tab:complexity} analyzes the computational cost of integrating Derain-Agent with Restormer at $256\times256$ resolution.
While the parameter count increases due to the inclusion of the decision network and refinement tools, the increase in FLOPs is marginal ($\sim$3.5\%).
This indicates that Derain-Agent is computationally efficient; the heavy lifting is done by the base derainer, while our agent performs lightweight, targeted corrections.
Consequently, Derain-Agent offers a favorable trade-off, delivering significant performance gains with minimal additional latency.

\begin{table}[t]
\centering
\caption{Model complexity analysis ($256\times256$ input). 
}\vspace{-2mm}
\label{tab:complexity}
\begin{adjustbox}{width=\linewidth}
\begin{tabular}{lccccc}
\toprule
\textbf{Method}  & \textbf{Params (M)} & \textbf{FLOPs (G)}  & \textbf{PSNR}& \textbf{SSIM}\\
\midrule
Restormer   & 26.09 & 140.9  & 31.59& 0.8835 \\
+ Derain-Agent  & 49.02 & 145.8  & 32.34 & 0.9074 \\

\bottomrule
\end{tabular}
\end{adjustbox}
\vspace{-2mm}
\end{table}

\subsection{Ablation Studies}
We conduct comprehensive ablation studies to validate the contribution of each component in Derain-Agent. All experiments utilize Restormer pretrained on Rain13K as the base derainer, with results reported on the LHP-Rain test set. 



\textbf{Execution Strategy.}
Table~\ref{execution} presents a comparative analysis of different execution strategies. Existing methods, such as \emph{IQA-Greedy} (from Q-Agent) and \emph{Rollback-Replanning} (from AgenticIR), rely on iterative search mechanisms—utilizing step-wise quality feedback or backtracking to rectify unfavorable steps, respectively. 
However, given a compact tool library, such exploration-based policies suffer from a restricted search space, failing to identify superior tool compositions and yielding only marginal improvements over the baseline. 
In contrast, \emph{Derain-Agent} learns to directly synthesize an effective tool program and modulates execution strength in a spatially adaptive manner. 
This approach delivers a substantial performance boost, achieving a \textbf{+0.75 dB} gain in PSNR and a significant reduction in LPIPS.



\begin{figure}[htbp]
  \centering 
  \includegraphics[width=\linewidth]{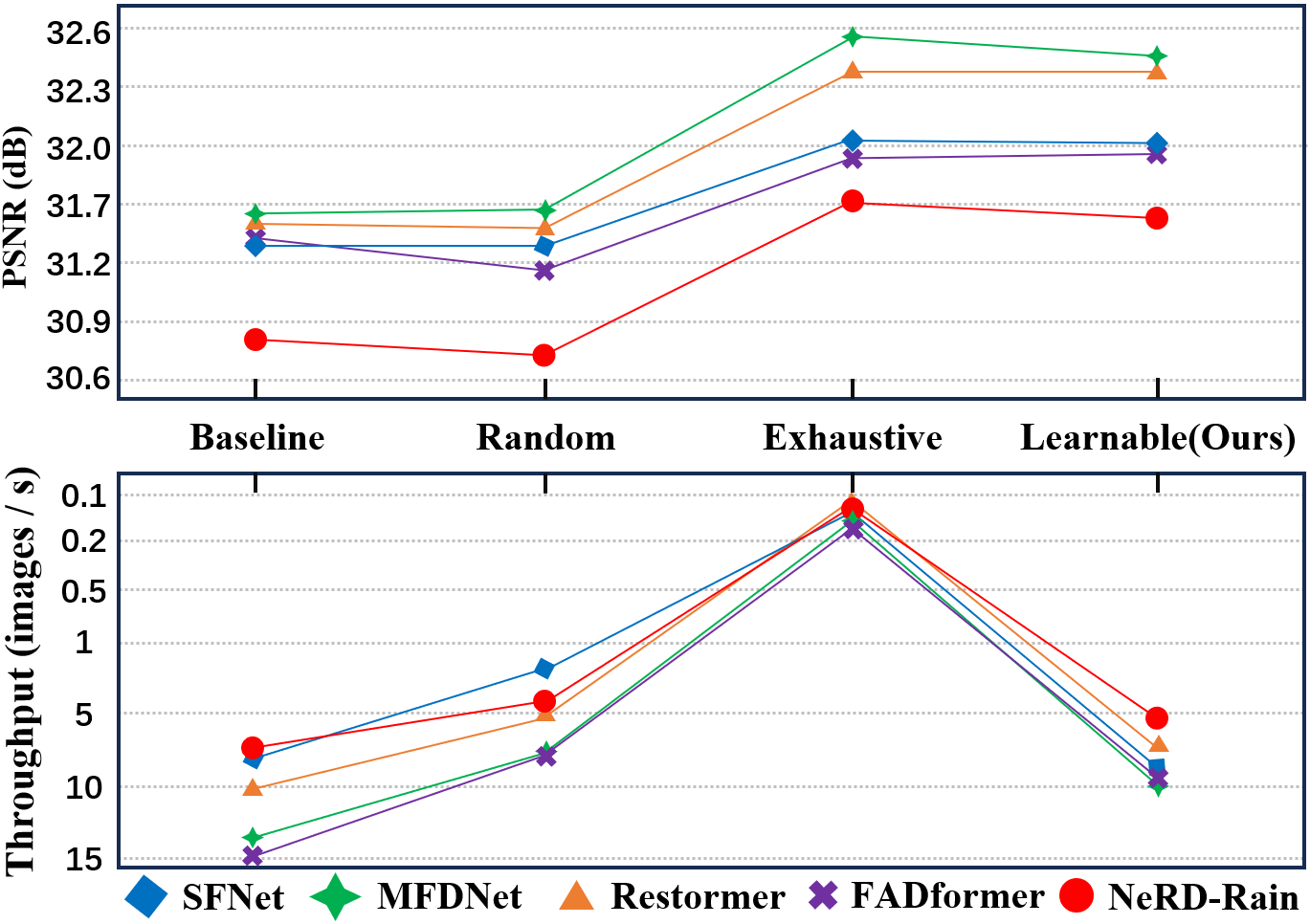}
  
  \caption{Performance–throughput trade-off of different tool-path scheduling strategies across different backbones. 
  }\vspace{-2mm}
  
  \label{path}
\end{figure}

\textbf{Tool Path Scheduling Strategies.}
We benchmark the proposed learnable scheduler against Baseline, Random, and Exhaustive strategies. 
As illustrated in Figure~\ref{path}, while \emph{Exhaustive search} establishes an upper bound in PSNR, it incurs a prohibitive computational cost, rendering it impractical for deployment.
\emph{Random scheduling} yields marginal and inconsistent improvements, indicating that arbitrary tool ordering may introduce adverse artifacts.
In contrast, our \emph{Learnable Scheduler} achieves near-oracle performance across various backbones while maintaining high throughput. This demonstrates that the policy network effectively amortizes the search cost, striking an optimal balance between restoration quality and inference efficiency.



\begin{table}[t]
\centering
\small
\caption{Ablation study on Execution Strategies under the same toolbox, backbone, and step budget. }\vspace{-2mm}
\label{execution}
\setlength{\tabcolsep}{2.3pt}
\begin{tabular}{lccc}
\toprule
\textbf{Strategy}  & \textbf{PSNR}  & \textbf{SSIM} & \textbf{LPIPS} 
\\
\midrule
Baseline   & 31.59 & 0.8835 & 0.219
\\
IQA-Greedy   & 31.65 & 0.8864 & 0.214
\\
Rollback-Replanning & 31.69 & 0.8873  & 0.215 
\\
Derain-Agent & 32.34  & 0.9074  & 0.177

\\

\bottomrule
\end{tabular}
\vspace{-2mm}
\end{table}


\begin{figure*}[!htpb]
  \centering
  \includegraphics[width=\textwidth]{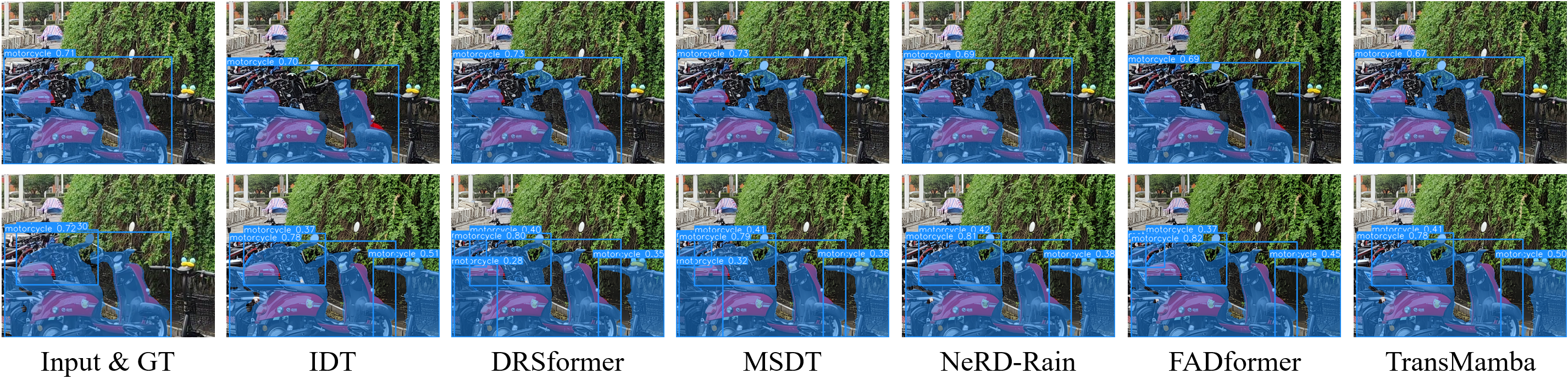}\vspace{-2mm}

  \caption{Visual comparison of object detection and instance segmentation on LHP-Rain. Top: Baseline derainers; Bottom: Derain-Agent enhanced. Our method significantly reduces false negatives and improves mask precision.}
  \label{down_task}\vspace{-4mm}
\end{figure*}

\begin{figure}[!htbp]
  \centering
  \includegraphics[width=\linewidth]{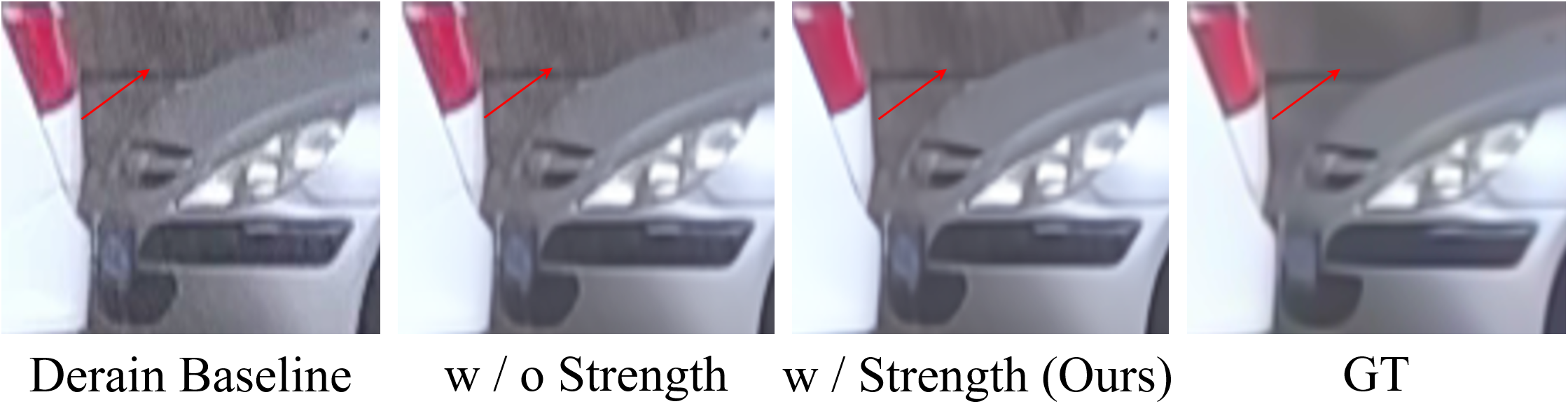}\vspace{-2mm}
  
  \caption{Qualitative comparison of local detail recovery. Strength modulation facilitates the reconstruction of fine spatial structures.}
  \label{strength}
  \vspace{-3mm}
\end{figure}
\textbf{Strength Modulation Mechanism.}
Table~\ref{tab-strength} quantifies the impact of the Strength Modulation (SM) mechanism. 
Incorporating SM yields a consistent gain of \textbf{0.09 dB} in PSNR on LHP-Rain with negligible overhead in parameters and FLOPs.
Qualitatively (see Figure~\ref{strength}), SM enables the agent to apply targeted refinement to regions with residual degradations, producing perceptually cleaner outputs with fewer artifacts compared to fixed-strength execution.
\begin{table}[t]
\centering
\small
\caption{Ablation study on strength modulation (SM). }\vspace{-2mm}
\label{tab-strength}
\setlength{\tabcolsep}{2.3pt}
\begin{tabular}{lcccc}
\toprule
\textbf{Strategy}  & \textbf{PSNR} 
& \textbf{SSIM}
& \textbf{Params (M)} & \textbf{FLOPs (G)} 
\\
\midrule
w / o SM   & 32.25 & 0.9037
& 21.42 & 4.804
\\
w / SM   & 32.34  & 0.9074 
& 22.93 & 4.901 
\\

\bottomrule
\end{tabular}
\vspace{-2mm}
\end{table}





\textbf{Toolbox Size.}
Table~\ref{train stages & Toolbox} analyzes the effect of toolbox diversity. Performance consistently improves as the toolbox expands from one to three tools. This trend suggests that the distinct tools (denoising, deblurring, and color correction) provide complementary benefits. A larger toolbox empowers the scheduling policy to compose more precise refinement programs tailored to specific residual degradations.

\begin{table}[t]
\centering
\small
\caption{Ablation study on training strategy and Toolbox Size.}\vspace{-2mm}
\label{train stages & Toolbox}
\begin{tabular}{lcccc}
\toprule
\textbf{Training Stages} & \textbf{One} & \textbf{Two} & \textbf{Three} &\\
\midrule
PSNR   & 32.04 & 32.34 & 32.16 \\
SSIM  & 0.8986  & 0.9074  & 0.9014\\
\midrule
\textbf{Toolbox Size} & \textbf{One} & \textbf{Two} & \textbf{Three} &\\
\midrule
PSNR   & 31.97  & 32.21 & 32.34 \\
SSIM  & 0.8925  &  0.9034 & 0.9074\\
\bottomrule
\end{tabular}
\vspace{-2mm}
\end{table}

\textbf{Training Strategy Analysis.}
Table~\ref{train stages & Toolbox} compares three training schemes.
\emph{One-stage} end-to-end training yields suboptimal results, as jointly optimizing discrete path selection and continuous strength regression leads to optimization conflicts.
Our proposed \emph{Two-stage} strategy achieves the best performance by decoupling these objectives: we first learn reliable tool paths and then regress execution strengths.
\emph{Third-stage} (full unfreezing) scheme slightly degrades performance, as re-coupling objectives disrupts the learned scheduling priors.

\begin{table}[t]
\centering
\small
\caption{Ablation study on different backbones.}\vspace{-2mm}
\label{tab-backbone}
\setlength{\tabcolsep}{2.3pt}
\begin{tabular}{lcccc}
\toprule
\textbf{Backbones}  & \textbf{PSNR} & \textbf{SSIM} & \textbf{Params (M)} & \textbf{FLOPs (G)} \\
\midrule

ResNet34   & 32.34 & 0.9074 & +22.93 & +4.901 \\
EfficientNet-B0     &  32.23&   0.9048& +11.63 & +2.372  \\
Shufflenetv2   & 32.24  & 0.9052  & +2.30 & +0.257  \\
Baseline   & 31.59 & 0.8835 & - & - \\

\bottomrule
\end{tabular}
\vspace{-2mm}
\end{table}


\textbf{Shared Backbones.}
Table~\ref{tab-backbone} evaluates the impact of the planning network's backbone. 
While ResNet34 achieves the highest performance, lighter architectures like EfficientNet-B0~\cite{tan2019efficientnet} and ShuffleNetV2~\cite{2018ShuffleNet} incur only minor performance drops ($\sim$0.1 dB).
This indicates that Derain-Agent is architecturally robust; the decision-making capability does not collapse with reduced capacity, facilitating deployment on resource-constrained devices.


\subsection{Impact on Downstream Task}
To assess the practical utility of Derain-Agent, we evaluate object detection and instance segmentation performance on the RE-Rain dataset using YOLOv8~\cite{UltralyticsYOLOv8}.
As shown in Figure~\ref{down_task}, images refined by Derain-Agent consistently yield higher detection accuracy and more precise instance masks.
These improvements stem from the agent's ability to adaptively mitigate task-interfering residuals (\emph{e.g.}, blur and noise) that persist after initial deraining. 
By suppressing degradations that disrupt object boundaries, Derain-Agent provides higher-fidelity inputs for downstream models, thereby enhancing perception reliability in complex rainy environments.

\section{Conclusion}

In this work, we propose Derain-Agent, a plug-and-play refinement framework that shifts single-image deraining from static inference to dynamic, agent-based planning. 
By predicting instance-specific restoration programs, comprising optimized tool sequences and spatially modulated strengths, Derain-Agent effectively addresses the complex, coupled degradations that limit conventional models.
Extensive experiments on both paired and real-world benchmarks demonstrate that our method consistently boosts the performance of diverse base derainers with only modest computational overhead. In future work, we aim to incorporate semantic priors from Large Language Models (LLMs) to enhance degradation diagnosis and to extend this adaptive scheduling paradigm to broader restoration challenges, such as dehazing and low-light enhancement.
\bibliographystyle{named}
\bibliography{ijcai26}

\begin{thebibliography}{}

\bibitem[\protect\citeauthoryear{Chen \bgroup \em et al.\egroup }{2021}]{chen2021pre}
Hanting Chen, Yunhe Wang, Tianyu Guo, Chang Xu, Yiping Deng, Zhenhua Liu, Siwei Ma, Chunjing Xu, Chao Xu, and Wen Gao.
\newblock Pre-trained image processing transformer.
\newblock In {\em Proceedings of the IEEE/CVF conference on computer vision and pattern recognition}, pages 12299--12310, 2021.

\bibitem[\protect\citeauthoryear{Chen \bgroup \em et al.\egroup }{2023}]{chen2023towards}
X~Chen, J~Pan, J~Dong, and J~Tang.
\newblock Towards unified deep image deraining: A survey and a new benchmark. arxiv.
\newblock {\em arXiv preprint arXiv:2310.03535}, 2023.

\bibitem[\protect\citeauthoryear{Chen \bgroup \em et al.\egroup }{2024a}]{restoreagent2024}
Haoyu Chen, Wenbo Li, Jinjin Gu, Jingjing Ren, Sixiang Chen, Tian Ye, Renjing Pei, Kaiwen Zhou, Fenglong Song, and Lei Zhu.
\newblock Restoreagent: Autonomous image restoration agent via multimodal large language models.
\newblock In {\em Advances in Neural Information Processing Systems (NeurIPS)}, 2024.

\bibitem[\protect\citeauthoryear{Chen \bgroup \em et al.\egroup }{2024b}]{chen2024bidirectional}
Xiang Chen, Jinshan Pan, and Jiangxin Dong.
\newblock Bidirectional multi-scale implicit neural representations for image deraining.
\newblock In {\em Proceedings of the IEEE/CVF Conference on Computer Vision and Pattern Recognition}, pages 25627--25636, 2024.

\bibitem[\protect\citeauthoryear{Cui \bgroup \em et al.\egroup }{2023}]{cui2023selective}
Yuning Cui, Yi~Tao, Zhenshan Bing, Wenqi Ren, Xinwei Gao, Xiaochun Cao, Kai Huang, and Alois Knoll.
\newblock Selective frequency network for image restoration.
\newblock In {\em The eleventh international conference on learning representations}, 2023.

\bibitem[\protect\citeauthoryear{Dang \bgroup \em et al.\egroup }{2023}]{dang2023efficient}
Jisheng Dang, Huicheng Zheng, Jinming Lai, Xu~Yan, and Yulan Guo.
\newblock Efficient and robust video object segmentation through isogenous memory sampling and frame relation mining.
\newblock {\em IEEE Transactions on Image Processing}, 32:3924--3938, 2023.

\bibitem[\protect\citeauthoryear{Dang \bgroup \em et al.\egroup }{2024}]{dang2024adaptive}
Jisheng Dang, Huicheng Zheng, Xiaohao Xu, Longguang Wang, Qingyong Hu, and Yulan Guo.
\newblock Adaptive sparse memory networks for efficient and robust video object segmentation.
\newblock {\em IEEE Transactions on Neural Networks and Learning Systems}, 2024.

\bibitem[\protect\citeauthoryear{Gao \bgroup \em et al.\egroup }{2024}]{gao2024efficient}
Ning Gao, Xingyu Jiang, Xiuhui Zhang, and Yue Deng.
\newblock Efficient frequency-domain image deraining with contrastive regularization.
\newblock In {\em European Conference on Computer Vision}, pages 240--257. Springer, 2024.

\bibitem[\protect\citeauthoryear{Guo \bgroup \em et al.\egroup }{2023}]{guo2023sky}
Yun Guo, Xueyao Xiao, Yi~Chang, Shumin Deng, and Luxin Yan.
\newblock From sky to the ground: A large-scale benchmark and simple baseline towards real rain removal.
\newblock In {\em Proceedings of the IEEE/CVF international conference on computer vision}, pages 12097--12107, 2023.

\bibitem[\protect\citeauthoryear{He \bgroup \em et al.\egroup }{2016}]{he2016deep}
Kaiming He, Xiangyu Zhang, Shaoqing Ren, and Jian Sun.
\newblock Deep residual learning for image recognition.
\newblock In {\em Proceedings of the IEEE conference on computer vision and pattern recognition}, pages 770--778, 2016.

\bibitem[\protect\citeauthoryear{Huynh-Thu and Ghanbari}{2008}]{huynh2008scope}
Quan Huynh-Thu and Mohammed Ghanbari.
\newblock Scope of validity of psnr in image/video quality assessment.
\newblock {\em Electronics letters}, 44(13):800--801, 2008.

\bibitem[\protect\citeauthoryear{Jiang \bgroup \em et al.\egroup }{2022a}]{jiang2022magic}
Kui Jiang, Zhongyuan Wang, Chen Chen, Zheng Wang, Laizhong Cui, and Chia-Wen Lin.
\newblock Magic elf: Image deraining meets association learning and transformer.
\newblock In {\em Proceedings of the 30th ACM International Conference on Multimedia}, pages 827--836, 2022.

\bibitem[\protect\citeauthoryear{Jiang \bgroup \em et al.\egroup }{2022b}]{jiang2022danet}
Kui Jiang, Zhongyuan Wang, Zheng Wang, Peng Yi, Junjun Jiang, Jinsheng Xiao, and Chia-Wen Lin.
\newblock Danet: Image deraining via dynamic association learning.
\newblock In {\em IJCAI}, pages 980--986, 2022.

\bibitem[\protect\citeauthoryear{Jiang \bgroup \em et al.\egroup }{2025}]{mair2025}
Xu~Jiang, Gehui Li, Bin Chen, and Jian Zhang.
\newblock Multi-agent image restoration.
\newblock {\em arXiv preprint arXiv:2503.09403}, 2025.

\bibitem[\protect\citeauthoryear{Jocher}{2023}]{UltralyticsYOLOv8}
Glenn Jocher.
\newblock Yolov8, 2023.
\newblock Version 8.x.

\bibitem[\protect\citeauthoryear{LeCun \bgroup \em et al.\egroup }{2015}]{lecun2015deep}
Yann LeCun, Yoshua Bengio, and Geoffrey Hinton.
\newblock Deep learning.
\newblock {\em nature}, 521(7553):436--444, 2015.

\bibitem[\protect\citeauthoryear{Li \bgroup \em et al.\egroup }{2024}]{li2024fouriermamba}
Dong Li, Yidi Liu, Xueyang Fu, Senyan Xu, and Zheng-Jun Zha.
\newblock Fouriermamba: Fourier learning integration with state space models for image deraining.
\newblock {\em arXiv preprint arXiv:2405.19450}, 2024.

\bibitem[\protect\citeauthoryear{Li \bgroup \em et al.\egroup }{2025}]{hybridagent2024}
Bingchen Li, Xin Li, Yiting Lu, and Zhibo Chen.
\newblock Hybrid agents for image restoration.
\newblock {\em arXiv preprint arXiv:2503.10120}, 2025.

\bibitem[\protect\citeauthoryear{Liu \bgroup \em et al.\egroup }{2024}]{liu2024vmamba}
Yue Liu, Yunjie Tian, Yuzhong Zhao, Hongtian Yu, Lingxi Xie, Yaowei Wang, Qixiang Ye, Jianbin Jiao, and Yunfan Liu.
\newblock Vmamba: Visual state space model.
\newblock {\em Advances in neural information processing systems}, 37:103031--103063, 2024.

\bibitem[\protect\citeauthoryear{Ma \bgroup \em et al.\egroup }{2018}]{2018ShuffleNet}
Ningning Ma, Xiangyu Zhang, Hai~Tao Zheng, and Jian Sun.
\newblock Shufflenet v2: Practical guidelines for efficient cnn architecture design.
\newblock {\em Springer, Cham}, 2018.

\bibitem[\protect\citeauthoryear{Manne and Wan}{2024}]{manne2024wb}
Sai Kumar~Reddy Manne and Michael Wan.
\newblock Wb luts: Contrastive learning for white balancing lookup tables.
\newblock {\em arXiv preprint arXiv:2404.10133}, 2024.

\bibitem[\protect\citeauthoryear{Mittal \bgroup \em et al.\egroup }{2012a}]{mittal2012no}
Anish Mittal, Anush~Krishna Moorthy, and Alan~Conrad Bovik.
\newblock No-reference image quality assessment in the spatial domain.
\newblock {\em IEEE Transactions on image processing}, 21(12):4695--4708, 2012.

\bibitem[\protect\citeauthoryear{Mittal \bgroup \em et al.\egroup }{2012b}]{mittal2012making}
Anish Mittal, Rajiv Soundararajan, and Alan~C Bovik.
\newblock Making a “completely blind” image quality analyzer.
\newblock {\em IEEE Signal processing letters}, 20(3):209--212, 2012.

\bibitem[\protect\citeauthoryear{Ren \bgroup \em et al.\egroup }{2019}]{ren2019progressive}
Dongwei Ren, Wangmeng Zuo, Qinghua Hu, Pengfei Zhu, and Deyu Meng.
\newblock Progressive image deraining networks: a better and simpler baseline.
\newblock In {\em CVPR}, pages 3937--3946, 2019.

\bibitem[\protect\citeauthoryear{Tan and Le}{2019}]{tan2019efficientnet}
Mingxing Tan and Quoc Le.
\newblock Efficientnet: Rethinking model scaling for convolutional neural networks.
\newblock In {\em International conference on machine learning}, pages 6105--6114. PMLR, 2019.

\bibitem[\protect\citeauthoryear{Wang \bgroup \em et al.\egroup }{2004}]{wang2004image}
Zhou Wang, Alan~C Bovik, Hamid~R Sheikh, and Eero~P Simoncelli.
\newblock Image quality assessment: from error visibility to structural similarity.
\newblock {\em IEEE transactions on image processing}, 13(4):600--612, 2004.

\bibitem[\protect\citeauthoryear{Wang \bgroup \em et al.\egroup }{2020}]{wang2020dcsfn}
Cong Wang, Xiaoying Xing, Yutong Wu, Zhixun Su, and Junyang Chen.
\newblock Dcsfn: Deep cross-scale fusion network for single image rain removal.
\newblock In {\em Proceedings of the 28th ACM international conference on multimedia}, pages 1643--1651, 2020.

\bibitem[\protect\citeauthoryear{Xiao \bgroup \em et al.\egroup }{2022}]{xiao2022image}
Jie Xiao, Xueyang Fu, Aiping Liu, Feng Wu, and Zheng-Jun Zha.
\newblock Image de-raining transformer.
\newblock {\em IEEE transactions on pattern analysis and machine intelligence}, 45(11):12978--12995, 2022.

\bibitem[\protect\citeauthoryear{Yang \bgroup \em et al.\egroup }{2017}]{yang2017deep}
Wenhan Yang, Robby~T Tan, Jiashi Feng, Jiaying Liu, Zongming Guo, and Shuicheng Yan.
\newblock Deep joint rain detection and removal from a single image.
\newblock In {\em Proceedings of the IEEE conference on computer vision and pattern recognition}, pages 1357--1366, 2017.

\bibitem[\protect\citeauthoryear{Zamir \bgroup \em et al.\egroup }{2021}]{zamir2021multi}
Syed~Waqas Zamir, Aditya Arora, Salman Khan, Munawar Hayat, Fahad~Shahbaz Khan, Ming-Hsuan Yang, and Ling Shao.
\newblock Multi-stage progressive image restoration.
\newblock In {\em Proceedings of the IEEE/CVF conference on computer vision and pattern recognition}, pages 14821--14831, 2021.

\bibitem[\protect\citeauthoryear{Zamir \bgroup \em et al.\egroup }{2022}]{zamir2022restormer}
Syed~Waqas Zamir, Aditya Arora, Salman Khan, Munawar Hayat, Fahad~Shahbaz Khan, and Ming-Hsuan Yang.
\newblock Restormer: Efficient transformer for high-resolution image restoration.
\newblock In {\em Proceedings of the IEEE/CVF conference on computer vision and pattern recognition}, pages 5728--5739, 2022.

\bibitem[\protect\citeauthoryear{Zhang \bgroup \em et al.\egroup }{2023}]{zhang2023practical}
Kai Zhang, Yawei Li, Jingyun Liang, Jiezhang Cao, Yulun Zhang, Hao Tang, Deng-Ping Fan, Radu Timofte, and Luc~Van Gool.
\newblock Practical blind image denoising via swin-conv-unet and data synthesis.
\newblock {\em Machine Intelligence Research}, 20(6):822--836, 2023.

\bibitem[\protect\citeauthoryear{Zhen \bgroup \em et al.\egroup }{2024}]{zhen2024freqmamba}
Zou Zhen, Yu~Hu, and Zhao Feng.
\newblock Freqmamba: Viewing mamba from a frequency perspective for image deraining.
\newblock {\em arXiv preprint arXiv:2404.09476}, 2024.

\bibitem[\protect\citeauthoryear{Zhou \bgroup \em et al.\egroup }{2025}]{zhou2025q}
Yingjie Zhou, Jiezhang Cao, Zicheng Zhang, Farong Wen, Yanwei Jiang, Jun Jia, Xiaohong Liu, Xiongkuo Min, and Guangtao Zhai.
\newblock Q-agent: Quality-driven chain-of-thought image restoration agent through robust multimodal large language model.
\newblock {\em arXiv preprint arXiv:2504.07148}, 2025.

\bibitem[\protect\citeauthoryear{Zhu \bgroup \em et al.\egroup }{2025}]{agenticir2025}
Kaiwen Zhu, Jinjin Gu, Zhiyuan You, Yu~Qiao, and Chao Dong.
\newblock An intelligent agentic system for complex image restoration problems.
\newblock In {\em International Conference on Learning Representations (ICLR)}, 2025.
\newblock Also available as arXiv:2410.17809.

\bibitem[\protect\citeauthoryear{Zuo \bgroup \em et al.\egroup }{2025}]{4kagent2025}
Yushen Zuo, Qi~Zheng, Mingyang Wu, Xinrui Jiang, Renjie Li, Jian Wang, Yide Zhang, Gengchen Mai, Lihong~V. Wang, James Zou, Xiaoyu Wang, Ming-Hsuan Yang, and Zhengzhong Tu.
\newblock 4kagent: Agentic any image to 4k super-resolution.
\newblock {\em arXiv preprint arXiv:2507.07105}, 2025.

\end{thebibliography}

\end{document}